\documentclass[11pt]{article}

\usepackage[margin=1in]{geometry}
\usepackage{comment}
\usepackage[utf8]{inputenc}
\usepackage[T1]{fontenc}
\usepackage{amsmath,amssymb}
\usepackage{mathpazo}
\usepackage{graphicx}
\usepackage{booktabs}
\usepackage{array}
\usepackage[usenames,dvipsnames]{xcolor}
\usepackage[colorlinks=true,linkcolor=blue,citecolor=blue,urlcolor=blue]{hyperref}
\usepackage{tabularx}
\usepackage{multirow}
\usepackage[most]{tcolorbox}
\usepackage{titlesec}
\usepackage{enumitem}
\usepackage{setspace}
\usepackage{parskip}

\hypersetup{pdftitle={The limits of exactness: On the failure of automatic differentiation in physics-informed machine learning},
            pdfauthor={Ameya D. Jagtap}}

\definecolor{Header}{HTML}{1F2A44}
\titleformat{\section}{\normalfont\Large\bfseries\color{Header}}{\thesection}{1em}{}
\titlespacing{\section}{0pt}{18pt}{8pt}

\newtcolorbox{keybox}[1]{
  colback=gray!8, colframe=Header, boxrule=0.4pt,
  leftrule=3pt, arc=1pt, outer arc=1pt,
  fonttitle=\bfseries\sffamily, title={#1},
  before skip=12pt, after skip=12pt
}

\begin{document}

\begin{flushleft}
{\sffamily\bfseries\small\color{Header} PERSPECTIVE}\par
\vspace{6pt}
{\Huge\bfseries The limits of exactness: On the failure of automatic differentiation in physics-informed machine learning}\par

\vspace{6pt}
Ameya D. Jagtap\footnote{Corresponding author: Ameya D. Jagtap (ajagtap@wpi.edu, ameyadjagtap@gmail.com)}\par

\textit{Aerospace Engineering Department, Worcester Polytechnic Institute, Worcester, MA 01609, USA}
\end{flushleft}

\vspace{10pt}
\noindent\textbf{\textit{
Automatic differentiation (AD) lets neural networks compute derivatives of governing equations to machine precision, and this precision has made it the computational backbone of physics-informed machine learning. Yet exactness in the mathematical sense is not the same as fidelity to the physics. Here I argue that a derivative can be numerically perfect and still be the wrong derivative for the problem at hand, because AD, by construction, has no notion of the physical structure a solution must obey. Convection and its associated directionality, diffusion, and dispersion are only the most visible instances of a much longer list that spans all branches of computational science and engineering, including conservation, thermodynamic consistency, symmetry, symplectic structure, positivity, monotonicity, and boundedness. Recognizing this broader gap reframes how the field should build the next generation of PDE-driven neural surrogates.
}}

\vspace{6pt}
\begin{small}Keywords: \textit{Automatic Differentiation}; \textit{Physics-Informed Machine Learning}; \textit{Neural PDE Surrogates}.
\end{small}

\vspace{4pt}
\hrule
\vspace{8pt}

\section*{Introduction}

Automatic differentiation (AD) is, in a real sense, the invention that made modern deep learning possible. Its reverse-mode variant, first described by Seppo Linnainmaa in the 1970s and later formalized by Andreas Griewank and others, became the mathematical engine behind backpropagation, and it now underlies almost every neural network trained today \cite{wengert1964simple,linnainmaa1976taylor,griewank2008evaluating,baydin2018automatic}. Over the past decade, AD has taken on a second job. Instead of only computing gradients of a loss function with respect to network weights, it is now routinely used to compute the derivatives that appear inside partial differential equations (PDEs) themselves; the ingredient at the heart of physics-informed neural networks (PINNs) \cite{raissi2019physics,jagtap2020extended}, SINDy \cite{brunton2016discovering}, the Deep Ritz method \cite{sirignano2018dgm}, the Deep Galerkin method \cite{yu2018deep}, physics-informed Kolmogorov-Arnold networks (PIKANs) \cite{shukla2024comprehensive,menon2026fekan}, physics-informed neural operators (PINOs) such as DeepONet \cite{wang2021learning} and Fourier Neural Operators \cite{li2021physics}, physics-informed generative models (PI-GEN) \cite{yang2020physics,mondal2026genvoid}, physics-informed computer vision (PICV) \cite{banerjee2024physics}, and scientific foundation models (SciFMs) \cite{menon2026scientific}.

This second job is seductive precisely because AD is exact. Unlike finite differences, AD carries no discretization error; it differentiates the network's computational graph term by term via the chain rule and returns a derivative accurate to floating-point precision. It is natural, then, to assume that a method built on exact derivatives will faithfully reproduce the physics encoded in a PDE. This paper argues that this assumption is only partly true, and that the gap between it and reality has practical consequences for how reliable PDE-driven neural surrogates actually are. The central question is simple to state: \emph{is mathematical exactness of a derivative sufficient to solve the problem it was computed for?} The answer is no, and the reason is more general than any single equation or application. AD is, definitionally, a pointwise, structure-agnostic operation: it returns the exact local slope of a function at a point, with no awareness of the global properties (conservation, symmetry, causality, admissibility) that the physical system the function represents is required to satisfy. Convection-dominated transport, where AD famously struggles \cite{penwarden2023unified,krishnapriyan2021characterizing}, is simply the most easily visualized member of a much larger family of such properties. This paper surveys that family, and asks what it means for AD to be reliable, rather than merely exact, across the physical sciences.

\section*{The reach of AD}
A first, often overlooked, limitation is definitional. Derivatives of a field $u$ with respect to a coordinate $\xi$ (space $x$ or time $t$) can be \emph{local}, meaning integer-order derivatives such as $\partial u/\partial \xi$ or $\partial^2 u/\partial \xi^2$, or \emph{nonlocal}, meaning fractional- or integral-order derivatives that depend on the entire history or domain of the function rather than on its behavior at a single point. AD, by its construction as a sequence of elementary local operations linked by the chain rule, can only ever reach the first kind. Fractional derivatives, which appear in anomalous diffusion, viscoelasticity, and other memory-dependent phenomena, require evaluating an integral quantity that has no finite chain-rule decomposition, so AD simply cannot be applied to them (see, Table~\ref{tab:Tab1}). This is a minor issue than the one this paper is mainly concerned with, but it is worth stating plainly: the reach of AD is bounded before questions of physical fidelity even arise.

\begin{table}[h]
\centering
\begin{tabular}{p{3.4cm} p{4.6cm} p{4.6cm}}
\toprule
\textbf{Derivative type} & \textbf{Example} & \textbf{Can AD evaluate it?} \\
\midrule
Local, integer-order & $\dfrac{\partial u}{\partial \xi},\ \dfrac{\partial^2 u}{\partial \xi^2},\ \ldots,\ \dfrac{\partial^n u}{\partial \xi^n}$ & Yes --- exactly, via the chain rule \\[8pt]
Nonlocal, fractional or integral-order & $\dfrac{\partial^{0.5} u}{\partial \xi^{0.5}},\ \dfrac{\partial^{-1} u}{\partial \xi^{-1}},\ \ldots$ & No --- requires evaluating an integral over the domain \\
\bottomrule
\end{tabular}
\caption{Local and nonlocal derivatives.}
\label{tab:Tab1}
\end{table}

\section*{When an exact derivative is still wrong}
Consider the linear convection equation, $u_t + \beta u_x = 0$, solved with a PINN. The network is trained by minimizing a composite loss,
\begin{equation}
\mathrm{Loss}(\theta) = w_d\, \mathrm{MSE}_d(\theta) + w_r\, \mathrm{MSE}_r(\theta),
\end{equation}
where $\mathrm{MSE}_d$ penalizes mismatch with known data and $\mathrm{MSE}_r$ penalizes the residual of the PDE itself, $\mathcal{F}(u) = u_t + \beta u_x$, evaluated using AD at collocation points. Because AD returns $u_t$ and $u_x$ to machine precision, the residual term is, numerically, exactly what it claims to be. And yet PINNs built this way are well known to struggle on convection-dominated problems, producing solutions that do not respect the direction the information in the true solution actually travels \cite{penwarden2023unified,krishnapriyan2021characterizing}.

The reason is that mathematical exactness of a derivative value says nothing about whether the numerical procedure built around it (using chain-rule) respects the physics of how information moves through the domain. AD evaluates $u_x$ as a single, direction-agnostic number at a point; it has no notion that, for a right-moving wave, the correct way to approximate that gradient numerically is to look upstream rather than symmetrically in both directions. Put simply, \textbf{AD does not respect the direction of information propagation}. This is not a flaw in AD's arithmetic  (the derivative it returns is correct in the calculus sense) but a mismatch between what exactness guarantees and what solving the PDE actually requires.

\vspace{0.2cm}
\noindent \textbf{A problem the numerical-methods community already solved once}: 
This tension is not new; it was simply solved in a different computational tradition. Richard Courant, Kurt Friedrichs, and Hans Lewy laid the groundwork for it in their foundational 1928 analysis of finite-difference stability \cite{courant1928partiellen}, work that eventually gave rise to \emph{upwind} methods: schemes that deliberately bias the derivative stencil toward the direction from which information physically arrives, rather than treating every direction symmetrically. Upwinding was formalized through the 1950s and 1960s and extended to shock-capturing flows by Jay Boris and David Book in the early 1970s \cite{boris1973flux}. More recent essentially non-oscillatory (ENO) \cite{harten1997uniformly} and weighted ENO (WENO) \cite{liu1994weighted} schemes refine the same idea, adapting the stencil itself near discontinuities and steep gradients so that the numerical solution tracks the true direction of propagation. Also, stabilized, locally conservative discontinuous Galerkin schemes rely on the same upwinding principle, using numerical fluxes at element interfaces to determine which neighboring cell supplies the information needed at that boundary, so that the discrete solution respects the true direction of propagation rather than mixing information from both sides indiscriminately \cite{cockburn2001runge,reed1973triangular}.

What these schemes share is that their design criterion is not mathematical exactness of a derivative at a point, but fidelity to the physical behavior of the system; stability, correct wave direction, and robustness near shocks. That criterion is largely orthogonal to what AD optimizes for. A PINN residual built purely from AD has no equivalent of \textit{upwind bias} built in; it must instead be inferred implicitly by network optimization, which explains why naive AD-based residuals are so often unstable on convection-dominated problems even though every derivative inside them is exact.

\section*{A broader catalogue of physical properties AD does not enforce}
\begin{figure}[htpb]
\centering
\includegraphics[scale=0.19, clip=true]{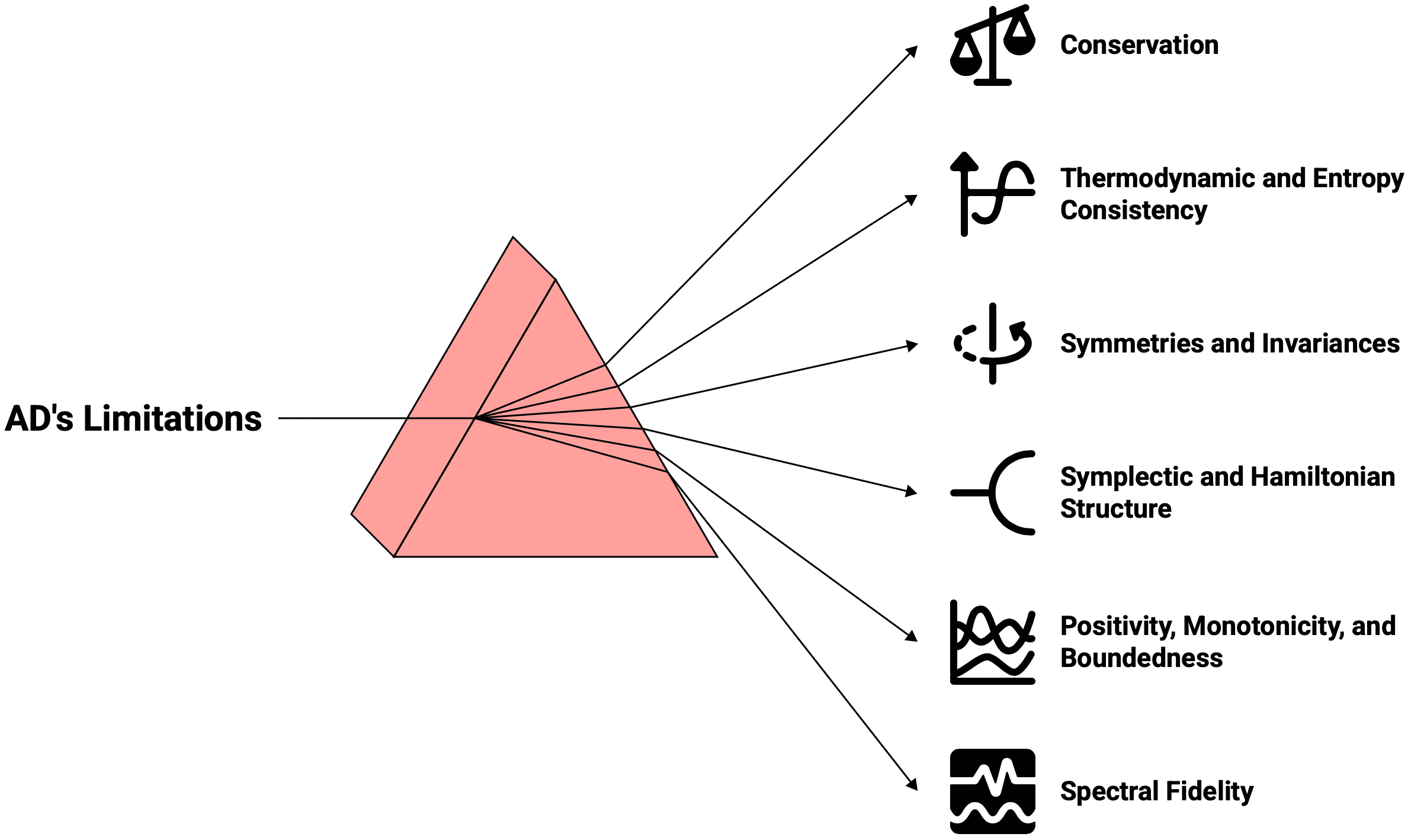}
\caption{Unveiling AD's Limitations in Physical Modeling}
\label{fig:ADFail}
\end{figure}

Directionality is the easiest member of this family to visualize, but it is one instance of a general fact: AD returns a number that is correct at a point, and correctness at a point is a much weaker statement than fidelity to the physics of the system as a whole. At least \textit{six} further properties recur across the physical sciences, none of which follows automatically from exact pointwise derivatives; see Figure~\ref{fig:ADFail} and Box 1.

\vspace{0.2cm}
\noindent \textbf{Conservation}: Many governing equations exist precisely because some quantity (mass, momentum, energy, charge) is conserved. A numerical or learned scheme can return locally exact derivatives of the governing fields while still failing to conserve the corresponding global quantity over long time integration, because conservation is a statement about the discrete or learned operator as a whole, not about any single derivative it contains. Kinetic-energy- and entropy-preserving finite-volume schemes for compressible flow were developed for exactly this reason \cite{chandrashekar2013kinetic,kuya2018kinetic}: they are constructed so that the numerical fluxes conserve kinetic energy and satisfy a discrete entropy inequality, a property that has nothing to do with how exactly any individual flux derivative is computed and everything to do with how the terms are assembled.

\vspace{0.2cm}
\noindent \textbf{Thermodynamic and entropy consistency}: Related to conservation, but distinct from it, is the requirement that irreversible processes not spontaneously decrease entropy. Physical solutions of dissipative PDEs satisfy a second-law-type inequality; a neural surrogate trained purely by minimizing an AD-based residual has no mechanism that prevents it from producing locally exact-looking derivatives that nonetheless correspond to an entropy-decreasing, and therefore unphysical, evolution. This is one reason entropy-stable and entropy-preserving formulations remain an active alternative to residual minimization alone \cite{chandrashekar2013kinetic,kuya2018kinetic}.

\vspace{0.2cm}
\noindent \textbf{Symmetries and invariances}: Physical laws typically hold regardless of translation, rotation, or the observer's frame of reference (Galilean invariance in classical mechanics, gauge invariance in electromagnetism, Lorentz invariance in relativistic field theories). A network trained with an AD-based residual can satisfy the PDE at every sampled collocation point and still fail to generalize a symmetry the true solution possesses, because nothing about pointwise derivative accuracy constrains how the learned function behaves under a transformation of its inputs. Respecting a symmetry is a property of the function's global structure, not of any derivative evaluated at a point. This is precisely why a separate line of work builds the symmetry into the architecture rather than hoping AD-based training will discover it: group-equivariant network layers guarantee equivariance to a specified transformation group by construction \cite{cohen2016group}, a principle that has been formalized for physical scalars, vectors, and tensors more broadly \cite{villar2021scalars} and demonstrated concretely by embedding Galilean and rotational invariance directly into turbulence-closure networks \cite{ling2016reynolds}.

\vspace{0.2cm}
\noindent \textbf{Symplectic and Hamiltonian structure}: Conservative dynamical systems (from molecular dynamics to celestial mechanics) evolve on a phase space equipped with a symplectic structure that constrains how volumes and energies behave over arbitrarily long integration times. Standard AD-based training respects neither property by default: a network can match a Hamiltonian system's equations of motion pointwise, with exact derivatives, while still drifting in total energy over long trajectories, because symplecticity is a property of the map from one time step to the next, not of the instantaneous derivative the map is built from. Hamiltonian neural networks address this by learning a scalar Hamiltonian and differentiating it through the canonical equations of motion rather than fitting the dynamics directly, which conserves energy by construction \cite{greydanus2019hamiltonian}; Lagrangian neural networks apply the same idea to systems more naturally expressed through a Lagrangian \cite{cranmer2020lagrangian}, and symplectic recurrent networks build the volume-preserving structure directly into the integrator used during training \cite{chen2019symplectic}.

\vspace{0.2cm}
\noindent \textbf{Positivity, monotonicity, and boundedness}: Many physical fields are constrained to a subset of the real line by their very definition: densities and pressures must stay positive, probabilities and volume fractions must stay between zero and one, concentrations cannot go negative, and solutions near a shock or phase boundary must remain monotone rather than oscillate. Maintaining these bounds discretely is itself a solved problem in numerical analysis: positivity-preserving limiters guarantee that density and pressure stay admissible in high-order finite-volume and discontinuous Galerkin schemes for compressible flow \cite{zhang2010maximum,zhang2010positivity}, while total-variation-diminishing (TVD) schemes and flux limiters enforce monotonicity near discontinuities so that no spurious oscillation drives a bounded quantity out of its physical range \cite{harten1997high,sweby1984high}. AD's exactness offers no guarantee of any of this. A network can return the exact derivative of a learned density field at every point while the field itself dips below zero between collocation points, because positivity, like conservation and symmetry, is a global constraint that pointwise exactness does not imply. Classical numerical analysis addressed this long before neural surrogates existed: flux-limiter schemes were designed specifically to suppress the spurious oscillations that unconstrained high-order discretizations produce near steep gradients \cite{sweby1984high}, and maximum-principle-satisfying, positivity-preserving finite-volume schemes guarantee bounded, non-negative solutions for systems such as compressible flow by construction \cite{zhang2010maximum}, again through a global constraint on the scheme rather than through any local derivative computation.

\vspace{0.2cm}
\noindent \textbf{Spectral fidelity: dissipation and dispersion}: Finally, the spectral behavior of a scheme (how it treats different frequency components of a solution) recurs across nearly every field that simulates wave-like phenomena. \emph{Dissipation}, associated with even-order derivatives $\partial^n u/\partial x^n$ for $n=2,4,6,\ldots$, damps wave amplitudes over time; some dissipation is often necessary for stability, but too much erases genuine physical detail. \emph{Dispersion}, associated with odd-order derivatives $\partial^n u/\partial x^n$ for $n=3,5,7,\ldots$, causes different frequency components to travel at different speeds, distorting wave shapes even when amplitude is preserved. Finite element analysis in solid mechanics \cite{hughes2003finite}, numerical heat transfer \cite{patankar2018numerical}, and finite-difference time-domain electromagnetics \cite{taflove2005computational} are all, in part, organized around controlling these two effects. AD offers no analogous control: it returns the exact value of $\partial^n u/\partial x^n$ wherever asked, but the learned solution as a whole can still be spuriously dissipative or dispersive, because nothing in an AD-based residual constrains the spectral behavior of the function it differentiates.

\begin{keybox}{Box 1 \textbar\ Six physical properties automatic differentiation does not enforce}
\begin{enumerate}[leftmargin=1.4em, itemsep=6pt, topsep=2pt]
\item \textbf{Conservation} --- locally exact derivatives can still fail to conserve mass, momentum, energy, or charge over long time integration. \textit{Domains}: compressible flow, electromagnetics, particle and rigid-body dynamics, plasma physics and MHD, climate and geophysical fluid dynamics, chemical reaction networks. \textit{Consequence if violated}: mass, momentum, energy, charge, or magnetic flux drift over long time integration.
\item \textbf{Thermodynamic / entropy consistency} --- nothing prevents an AD-based residual from producing an entropy-decreasing, unphysical evolution. \textit{Domains}: compressible and dissipative flows, heat transfer, combustion, phase-field and materials modeling. \textit{Consequence if violated}: locally exact but entropy-decreasing, unphysical evolution.
\item \textbf{Symmetries and invariances} --- pointwise accuracy at collocation points does not constrain how a learned function behaves under translation, rotation, or a change of frame. \textit{Domains}: classical and relativistic field theories, electromagnetism, quantum chemistry, molecular simulation. \textit{Consequence if violated}: poor generalization across translated, rotated, or boosted inputs.
\item \textbf{Symplectic / Hamiltonian structure} --- a network can match a Hamiltonian system's equations of motion pointwise while still drifting in total energy over long trajectories. \textit{Domains}: molecular dynamics, celestial mechanics, rigid-body systems, plasma and beam dynamics. \textit{Consequence if violated}: spurious energy or phase-space-volume drift over long trajectories.
\item \textbf{Positivity, monotonicity, and boundedness} --- densities, probabilities, and concentrations can dip outside their admissible range between collocation points even where every sampled derivative is exact. \textit{Domains}: density and pressure fields, probabilities, shock and phase-boundary regions, chemical kinetics, population and epidemiological models. \textit{Consequence if violated}: negative densities or concentrations, oscillatory or non-physical solutions.
\item \textbf{Spectral fidelity (dissipation and dispersion)} --- AD returns the exact value of any derivative order asked for, but nothing constrains the resulting solution's damping or phase behaviour across frequency components. \textit{Domains}: solid mechanics, acoustics, electromagnetics, heat transfer, seismology, numerical relativity. \textit{Consequence if violated}: excessive damping, phase errors, or constraint violation in wave-like solutions.
\end{enumerate}
\end{keybox}

\section*{Why this cuts across domains}
\begin{figure}[htpb]
\centering
\includegraphics[scale=0.55, clip=true]{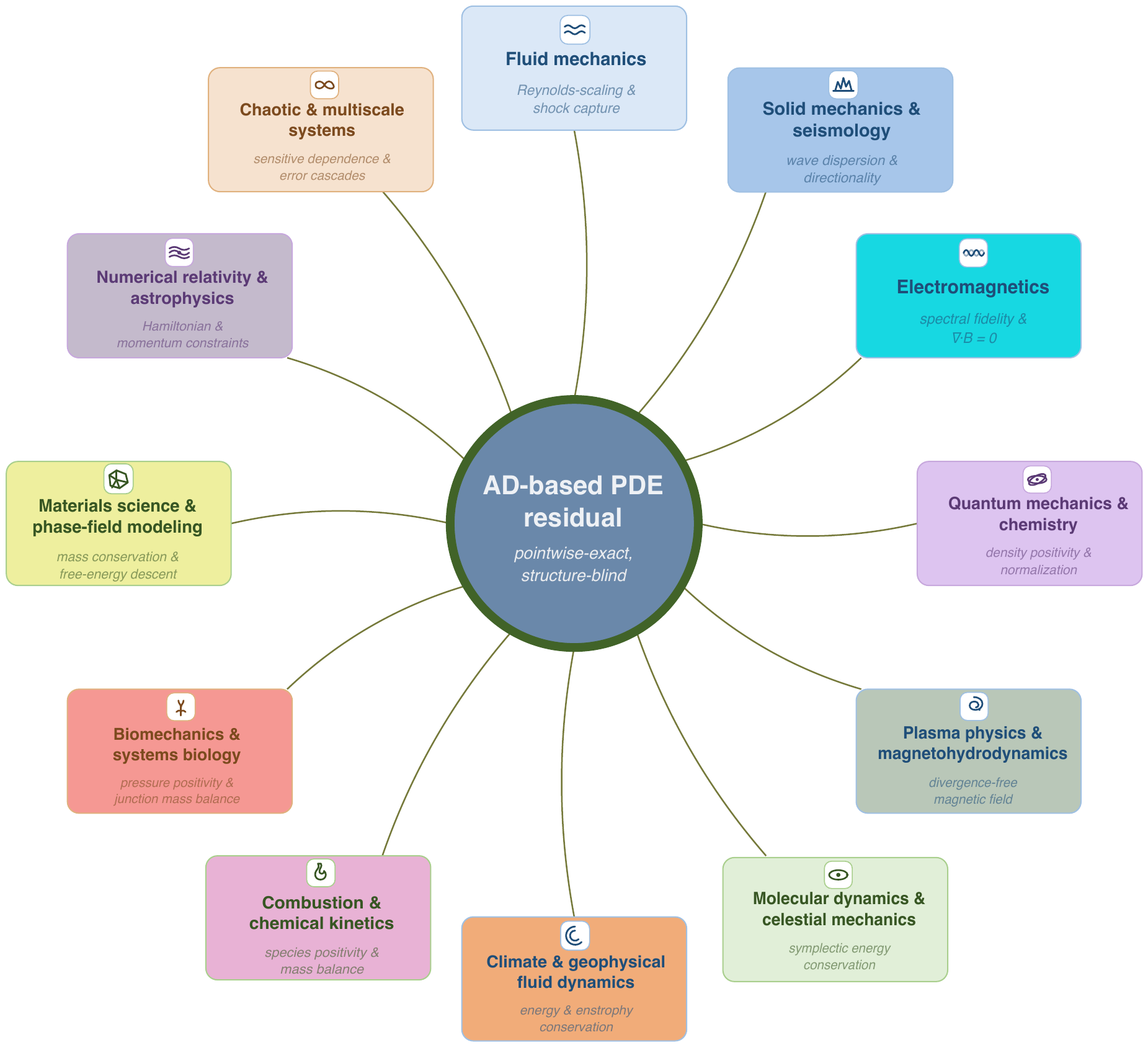}
\caption{Structural Blind Spots of Automatic Differentiation in Physics-Informed Machine Learning Across Computational Science and Engineering}
\label{fig:ADIsu}
\end{figure}
\noindent \textbf{Fluid mechanics}: Convection-dominated flow remains the most thoroughly documented example, and it is worth summarizing briefly here precisely because it shows the general pattern so clearly. Lid-driven cavity flows governed by the incompressible Navier--Stokes equations 
are a standard benchmark on which vanilla AD-based PINNs, and several reweighting variants developed to stabilize them, show accuracy that degrades as the Reynolds number rises, matching classical solutions \cite{ghia1982high} well at $Re=100$ but increasingly poorly by $Re=1000$ and $Re=5000$; adding a parameterized artificial eddy viscosity term to the residual extends agreement with direct numerical simulation to $Re=2000$--$5000$, but the underlying difficulty at high Reynolds number persists \cite{wang2023solution}. Discontinuous solutions of hyperbolic conservation laws pose a related but distinct challenge, because AD-based residuals are smooth by construction and cannot represent a shock; successful applications to compressible, shock-containing flows have so far leaned on inverse-problem settings where additional data compensates for what the residual alone cannot capture \cite{mao2020physics,jagtap2022physics}.

\vspace{0.2cm}
\noindent \textbf{Solid mechanics and seismology}: Wave propagation through an elastic medium, as in earthquake engineering, ultrasonic non-destructive testing, or seismic imaging; depends on the same dissipation and dispersion control that Hughes's treatment of finite elements addresses \cite{hughes2003finite}, independent of any fluid convection term. Velocity--stress finite-difference formulations for seismic wave propagation were specifically designed so that stress and velocity are staggered in space and time to preserve the correct wave-propagation direction and avoid the numerical dispersion that a naive, symmetric discretization introduces \cite{virieux1986p}; an AD-based residual evaluated on a learned displacement field offers no analogous control over how the different frequency components of a seismic wavefield disperse.

\vspace{0.2cm}
\noindent \textbf{Electromagnetics}: In electromagnetics, radar cross-section analysis and antenna design depend on preserving spectral fidelity in Maxwell's equations under finite-difference time-domain discretization \cite{taflove2005computational}, a wave-propagation problem with no convective term at all. Maxwell's equations additionally carry a structural constraint, $\nabla\cdot\mathbf{B}=0$, that has nothing to do with the accuracy of any individual spatial derivative: numerical schemes that do not enforce it can accumulate a spurious, unphysical magnetic monopole density over the course of a simulation, a failure mode documented in the plasma and electromagnetics literature well before neural surrogates existed \cite{brackbill1980effect} and one that an AD-based residual, which evaluates each field derivative independently, does nothing to prevent.

\vspace{0.2cm}
\noindent \textbf{Quantum mechanics and quantum chemistry}: In quantum mechanics, a learned wavefunction or density can satisfy the Schr\"odinger equation pointwise while violating the positivity or normalization that probability interpretation requires. Electronic-structure methods built on the \textit{Kohn--Sham formulation} of density functional theory manage this by construction, expressing the electron density through a set of orbitals whose occupation and normalization are fixed by the theory rather than left to an unconstrained residual \cite{kohn1965self}; a neural surrogate that instead regresses the density directly against an AD-based Schr\"odinger residual has no equivalent safeguard against a locally exact but negative or non-normalized density.

\vspace{0.2cm}
\noindent \textbf{Plasma physics and magnetohydrodynamics}: Magnetohydrodynamic (MHD) simulations of fusion and astrophysical plasmas inherit both the directionality and conservation issues seen in fluid dynamics, plus the divergence-free constraint on the magnetic field noted above. Brackbill and Barnes showed that failing to enforce $\nabla\cdot\mathbf{B}=0$ numerically introduces a spurious force parallel to the field lines that has no physical counterpart \cite{brackbill1980effect}; the fix is a property of how the discrete operator is assembled, not of the exactness of any single field derivative, and an AD-based residual built directly from Maxwell's and the momentum equations does not supply it automatically.

\vspace{0.2cm}
\noindent \textbf{Molecular dynamics and celestial mechanics}: In molecular dynamics and celestial mechanics, energy and phase-space volume must be preserved over astronomically long integration windows, which is a symplectic-structure requirement rather than a directionality or spectral one. Symplectic integrators for the $n$-body problem were developed specifically because standard, non-symplectic time-stepping schemes accumulate a secular energy drift over long integrations even though each individual force evaluation is exact \cite{wisdom1991symplectic}; the same logic motivates Hamiltonian and Lagrangian neural networks, discussed above, as the learned-surrogate analogue of this classical fix \cite{greydanus2019hamiltonian,cranmer2020lagrangian}.

\vspace{0.2cm}
\noindent \textbf{Climate and geophysical fluid dynamics}: Long-duration climate and ocean simulations depend on numerical schemes that conserve integral quantities such as \textit{total energy} and \textit{enstrophy} over thousands of simulated time steps, a property that is invisible at the level of any single derivative evaluation. Arakawa's finite-difference formulation of two-dimensional incompressible flow was constructed precisely to conserve both quantities simultaneously, a design goal orthogonal to the pointwise accuracy of the derivatives involved \cite{arakawa1997computational}; a neural surrogate trained by minimizing an AD-based residual at scattered collocation points has no comparable guarantee and can drift in exactly the integral quantities a climate model most needs to preserve.

\vspace{0.2cm}
\noindent \textbf{Combustion and chemical kinetics}: Combustion and reacting-flow simulations couple stiff chemical kinetics to the flow equations, and the species concentrations and reaction rates involved must remain non-negative and, for closed reaction networks, satisfy overall mass balance. Reduced and stiff chemical-kinetic mechanisms are built specifically to preserve these structural properties under aggressive time-stepping \cite{lu2009toward}; an AD-based residual differentiating a learned species field pointwise offers no corresponding guarantee that the learned concentrations stay non-negative or that elemental mass is conserved across reactions.

\vspace{0.2cm}
\noindent \textbf{Biomechanics and systems biology}: Cardiovascular and other physiological flow models add positivity of pressure and flow rate, together with mass conservation at vessel junctions, to the list of properties a numerical or learned scheme must respect; these one-dimensional and reduced-order circulatory models are constructed explicitly around conserving mass and momentum at branching points \cite{formaggia2010cardiovascular}. An AD-based PINN residual evaluated along a vessel network enforces the local governing equations pointwise but has no built-in mechanism for guaranteeing that pressure remains positive or that flow is conserved at every junction simultaneously.

\vspace{0.2cm}
\noindent \textbf{Materials science and phase-field modeling}: Phase-field models of microstructural evolution, such as the Cahn--Hilliard equation for phase separation, are fourth-order in space precisely because they are built to conserve the total amount of the diffusing species while allowing free energy to decrease monotonically \cite{cahn1958free}. Both properties are global, integral statements about the solution; an AD-based residual can return an exact value of the fourth-order spatial derivative at a point while the learned field still fails to conserve mass or violates the monotone decrease of free energy over the simulated evolution.

\vspace{0.2cm}
\noindent \textbf{Numerical relativity and astrophysics}: Numerical relativity evolves Einstein's field equations subject to Hamiltonian and momentum constraints that the true solution must satisfy at every instant; free-evolution schemes are explicitly designed to control the growth of constraint violations over long simulations, since generic finite-difference or spectral discretizations otherwise allow these violations to grow without bound even when each metric derivative is computed accurately \cite{baumgarte2010numerical}. A neural surrogate trained on an AD-based residual of the field equations inherits exactly this risk: it has no intrinsic mechanism for keeping the constraints satisfied over the course of a long evolution.

\vspace{0.2cm}
\noindent \textbf{Chaotic and multiscale systems}: Finally, in chaotic or multiscale systems more broadly (from turbulence to weather prediction) small pointwise derivative errors, however exact in isolation, can be amplified by sensitive dependence on initial conditions regardless of the physical domain. This sensitivity is quantified directly through the Lyapunov exponents of the underlying dynamics \cite{wolf1985determining}, and in multiscale flows it manifests as an error cascade in which small-scale inaccuracies propagate to progressively larger scales \cite{lorenz1969predictability}, mirroring the energy cascade that governs turbulence itself \cite{kolmogorov1941local} and setting a practical limit on predictability in weather and climate forecasting \cite{palmer2000predicting}. 

Across every discipline surveyed here, the failure mode is the same: AD supplies an exact local derivative, and the property that fails is a global one (conservation, a divergence-free constraint, positivity, symplecticity, or a bounded constraint violation) that the training procedure never explicitly enforced; see Figure~\ref{fig:ADIsu}.

\section*{How the field is responding}
None of this argues for abandoning AD in scientific machine learning. Rather, it argues that AD needs the same design discipline long applied to finite-difference, finite-volume and finite-element schemes, discipline that must now be tailored to whichever physical property is at stake (Table~\ref{tab:strategies}).

\paragraph{Restoring directionality:} Four strategies have taken hold. The most direct simply adds a small, explicit diffusion term to a hyperbolic residual, mimicking a central-difference scheme and restoring a preferred direction of information flow \cite{fuks2020limitations}. A second approach, curriculum learning, starts training with an artificially inflated diffusion coefficient and gradually anneals it toward its true, often near-vanishing, value, keeping the loss landscape navigable until the convection-dominated regime is reached \cite{krishnapriyan2021characterizing}. A third decomposes the domain in time, transferring the learned solution across subdomains so that causal ordering is imposed by construction rather than left to chance \cite{penwarden2023unified}. A fourth reweights the loss itself, through a single global scalar \cite{wang2022and}, pointwise and self-adaptive weights \cite{mcclenny2023self}, or by redirecting individual loss gradients \cite{menon2026tackling}, to keep data and residual terms in comparable proportion as training proceeds. A more radical line of work dispenses with AD's derivatives altogether in favour of direction- and stencil-aware numerical differentiation: CAN-PINN couples numerical and automatic differentiation explicitly, using the former to inject the correct upwind bias \cite{chiu2022can}, while DT-PINN shows that numerical derivatives can be simultaneously more physically faithful and computationally cheaper than AD when directionality matters \cite{sharma2022accelerated}. At the level of the full simulation loop, neural operators have been coupled directly to a high-fidelity/direct numerical simulation solver, using the learned operator to accelerate or bypass expensive steps while the DNS solver periodically corrects or re-initializes the trajectory, blending data-driven speed with numerical-solver accuracy over long rollouts \cite{oommen2022learning}. A closely related hybrid pairs a Fourier neural operator with a classical U-Net-style correction network to keep phase-field predictions numerically consistent over long horizons \cite{bonneville2025accelerating}.

\paragraph{Restoring conservation and thermodynamic consistency:} Entropy-stable and kinetic-energy-preserving flux formulations, developed originally for classical finite-volume discretizations, have been built directly into the loss or the network architecture \cite{jagtap2022physics,patel2022thermodynamically}. This idea now extends to fully structure-preserving PINN formulations that enforce energy dissipation or free-energy decrease for phase-field equations such as Allen--Cahn \cite{kutuk2025energy}.

\paragraph{Restoring invariance:} Here the response has split into loss-based and architecture-based routes. On the loss side, the Lie-point and generalized symmetries of a governing PDE can be added as an invariant-surface-condition penalty, either directly in the PINN loss \cite{arora2024invariant} or as a symmetry loss that lets a DeepONet learn one member of a solution family and infer the rest by transformation \cite{akhound2023lie}. A related strategy filters the candidate library in equation discovery down to Galilean- or Lorentz-invariant terms before fitting \cite{chen2024invariance}, and Lie-symmetry-informed domain decomposition merges this idea with the same time-subdomain logic used for directionality \cite{liu2024symmetry}. On the architecture side, constraints are imposed by construction rather than penalty: DeepONets can be built to satisfy boundary and symmetry constraints exactly \cite{brecht2023improving}, finite-group-equivariant layers extend a network's weight structure so that a discrete symmetry holds at every layer \cite{liu2024symmetry}, and lattice-based physical symmetries can be encoded directly into the network's functional form \cite{zhu2022neural}. A newer, more general strategy folds symmetry reduction into the solution ansatz itself, using an auxiliary network to learn the time-dependent transformation that renders a symmetry-driven solution stationary in rescaled coordinates \cite{kavousanakis2026going}.

\paragraph{Restoring symplecticity:} The guiding principle here is that a time-stepping scheme, not AD, should carry the burden of preserving phase-space structure. SympNets and their successors parameterize the flow map itself as a composition of exactly symplectic layers, rather than differentiating a learned Hamiltonian and hoping the resulting vector field integrates well \cite{jin2020sympnets,he2024deep}. The idea has since diversified: pseudo-symplectic networks trade a small, controlled violation of exact symplecticity for better data efficiency \cite{cheng2025learning}; hybrid schemes learn the Hamiltonian with a physics-informed network and hand the time evolution to a classical high-order symplectic integrator, so structure preservation is inherited rather than learned \cite{liang2025spini}; the same construction has been lifted from finite-dimensional Hamiltonian systems to infinite-dimensional Hamiltonian PDEs via symplectic neural operators \cite{makara2026symplectic}; and long-time stability for non-canonical systems has been addressed through Poisson-structure-preserving networks \cite{courtes2025neural}, with generalized Hamiltonian formulations following the same architectural logic \cite{makara2026symplectic}.

\paragraph{Restoring positivity, monotonicity and boundedness:} The common thread is that the output layer, or an explicit projection step, should make violation structurally impossible, not merely penalized. Hard-constrained output layers, first introduced for inverse design problems, reparameterize the network output so that boundary and box constraints hold exactly rather than approximately \cite{lu2021physics}, an approach that can be combined with exact distance-function constructions satisfying boundary conditions identically \cite{sukumar2022exact}. Structure-preserving PINNs built for the Allen--Cahn and related phase-field equations enforce boundedness of the order parameter and monotone energy decay by construction rather than by loss penalty \cite{kutuk2025energy}, and weak-adversarial-network formulations recast the constrained optimization problem itself, using a primal--dual game to keep the solution inside the feasible set throughout training rather than only at convergence \cite{bao2024wanco}. Positivity can also be enforced at the structural level, as shown for high-speed flows \cite{jagtap2022physics}.

Taken together, these efforts share a single premise: none asks AD to do a job it was never built for. Each preserves AD's exactness where it is genuinely load-bearing, differentiating the network with respect to its own parameters, while handing the physical structure of the solution to a scheme, constraint or architectural choice designed specifically to respect it.

\begin{table}[htpb]
\centering
\small
\begin{tabularx}{\textwidth}{@{} l X X @{}}
\toprule
\textbf{Property} & \textbf{Representative strategies} & \textbf{Main trade-off} \\
\midrule
Directionality &
Artificial diffusion \cite{fuks2020limitations}; curriculum learning \cite{krishnapriyan2021characterizing}; time-domain decomposition \cite{penwarden2023unified}; scalar or self-adaptive loss reweighting \cite{wang2022and,mcclenny2023self}; hybrid numerical--AD stencils \cite{chiu2022can,sharma2022accelerated}; hybrid solver 
\cite{oommen2022learning,bonneville2025accelerating} &
Each restores an upwind bias or causal ordering by hand-tuned terms, schedules, or an explicit mesh, reintroducing the discretization choices AD was meant to avoid. \\
\addlinespace
Conservation / thermodynamic consistency &
Entropy-stable / kinetic-energy-preserving fluxes built into the loss or architecture \cite{chandrashekar2013kinetic,kuya2018kinetic}; structure-preserving PINNs enforcing energy dissipation \cite{kutuk2025energy} &
Requires a known discrete flux or free-energy form; demonstrated mainly on phase-field-type equations so far. \\
\addlinespace
Invariance &
Symmetry-loss penalties and invariant-surface conditions \cite{arora2024invariant,akhound2023lie,chen2024invariance,liu2024symmetry}; hard-constrained or equivariant architectures \cite{brecht2023improving,zhu2022neural}; symmetry-reduced ansatz and operators \cite{kavousanakis2026going} &
Loss-based methods enforce symmetry only at sampled points; architectural methods are limited to symmetries with a known algebraic representation. \\
\addlinespace
Symplecticity &
Symplectic-by-construction networks \cite{he2024deep} and pseudo-symplectic relaxations \cite{cheng2025learning}; hybrid PINN--symplectic integrators \cite{liang2025spini}; symplectic neural operators \cite{makara2026symplectic}; non-canonical / nearly-periodic structure-preserving networks \cite{courtes2025neural} &
Restricts the hypothesis class to (near-)symplectic maps, or shifts the structure-preservation burden onto a classical integrator whose accuracy depends on the learned Hamiltonian. \\
\addlinespace
Positivity / monotonicity / boundedness &
Hard-constrained output layers \cite{lu2021physics,sukumar2022exact}; structure-preserving PINNs for bounded order parameters \cite{kutuk2025energy}; constrained-optimization / adversarial reformulation \cite{bao2024wanco} &
Requires the constraint form to be known analytically in advance, or introduces the optimization instability of adversarial training. \\
\bottomrule
\end{tabularx}
\caption{Strategies for restoring physical fidelity across the properties in Box 1.}
\label{tab:strategies}
\end{table}

\vspace{-0.2cm}
\section*{Conclusions}
Automatic differentiation (AD) in physics-informed machine learning framework promised something seductive: that the hard-won machinery of numerical analysis (the meshes, the limiters, the carefully constructed schemes) could be bypassed. Write down a residual, differentiate it exactly, and let the network find its way to a solution. What this perspective has shown, case after case, is that this promise is only half true. \textit{AD is exact about arithmetic; it is silent about physics}. Directionality made this vivid: a scheme can differentiate a discontinuous solution perfectly and still smear it, because exactness in computing a derivative says nothing about whether information is allowed to travel upwind or downwind. The same gap reappears under different names throughout the physical sciences; a network can conserve energy on the training set and leak it over long integration, respect a symmetry in-distribution and break it under extrapolation, stay positive on every example it was shown and go negative on the next. In each case, the flaw is not that AD computed the wrong derivative. It computed the right one. The problem is that exactness of differentiation and fidelity to physical law are simply different properties, governed by different mathematics, and nothing about AD guarantees the second just because it delivers the first.
This is why blind reliance on AD is dangerous. Treating its exactness as a proxy for physical correctness is not a minor approximation but a category error; mistaking a property of arithmetic for a property of nature. None of this is an argument against AD itself, only against using it unconditionally. The corrective is not a bigger network or a better-tuned loss, but the same discipline numerical analysis has practiced for a century: identify which structural property (conservation, symmetry, boundedness, symplectic structure) actually governs whether a solution can be trusted, and enforce it deliberately, by mesh, by constraint, or by architecture, rather than hoping it emerges. That costs something: the simplicity and generality that made AD-based methods appealing in the first place. But it is a cost worth paying selectively, wherever physical fidelity is the thing actually at stake; because a surrogate that is merely accurate is not the same as a surrogate that is reliable.

In summary, this perspective points toward five conclusions.
\begin{itemize}
\item \textbf{Exactness is not fidelity}: A derivative computed by AD is correct in the calculus sense at every point it is evaluated, but this pointwise correctness carries no guarantee that the physical structure of the solution, such as, its directionality, conservation, symmetry, symplecticity, or boundedness, is respected globally.

\item \textbf{The gap is general}: Convection-dominated flow is the most visible and thoroughly documented failure mode, but the same structural gap recurs, in different guises, across essentially all computational science and engineering fields.

\item \textbf{Each property demands its own fix}: Directionality, conservation, thermodynamic consistency, symmetry, symplectic structure, and positivity are mathematically distinct requirements, and the numerical-methods and structure-preserving machine-learning literature already offers targeted solutions for each; no single architectural or loss-based trick addresses all of them at once.

\item \textbf{Standard benchmarking is an incomplete test of reliability}: Accuracy on a held-out test set does not establish that a surrogate conserves the quantities it should, stays within physically admissible bounds, or generalizes a symmetry, and these structural properties need their own, explicitly designed diagnostics.

\item \textbf{The path forward is selective hybridization}: AD should be retained where its exactness is genuinely load-bearing (differentiating the network with respect to its own parameters) while the physical structure of the solution itself should be delegated to whichever classical scheme, constraint, or architecture was built to preserve the specific property the application actually depends on.
\end{itemize}

\small
\bibliographystyle{unsrt}
\bibliography{refs}
\end{document}